\pdfoutput=1
\documentclass[10pt, a4paper]{article}

\usepackage[final]{lrec2026}

\usepackage{times}
\usepackage{latexsym}

\usepackage[T1]{fontenc}
\usepackage[utf8]{inputenc}

\usepackage{microtype}

\usepackage{inconsolata}

\usepackage{graphicx}
\usepackage{inconsolata}
\usepackage{booktabs}
\usepackage{amsmath} 
\usepackage{tabularx}
\usepackage{arydshln}
\usepackage{enumitem}
\usepackage{comment}

\usepackage{algorithmic,algorithm}
\renewcommand{\algorithmiccomment}[1]{\bgroup\hfill$\triangleright$~#1\egroup}
\usepackage[linguistics]{forest} 
\usepackage{synttree}
\usepackage{tikz-dependency}
\usepackage{caption}
\usepackage{subcaption}
\usepackage{tablefootnote}

\usepackage{arydshln}

\usepackage{langsci-gb4e} 

\usepackage{listings}
\usepackage{xcolor}

\usepackage{siunitx}

\usepackage{orcidlink}

\title{Can LLMs Simulate Human Behavioral Variability? A Case Study in the Phonemic Fluency Task}

\name{
{Mengyang Qiu\textsuperscript{1}}\orcidlink{0000-0002-9398-7080}~~~{Zoe Brisebois\textsuperscript{2}}~~~{Siena Sun\textsuperscript{1}}
} 

\address{\textsuperscript{1}Department of Speech-Language Pathology, Saint Elizabeth University, United States \\
         \textsuperscript{2}Department of Psychology, Trent University, Canada \\
         \textsuperscript{1}{\tt \{mqiu, ssun\}@steu.edu}~~~\textsuperscript{2}{\tt zoebrisebois@trentu.ca}\\}

\abstract{
Large language models (LLMs) are increasingly explored as substitutes for human participants in cognitive tasks, but their ability to simulate human behavioral variability remains unclear. This study examines whether LLMs can approximate individual differences in the phonemic fluency task, where participants generate words beginning with a target letter. We evaluated 34 distinct models across 45 configurations from major closed-source and open-source providers, and compared outputs to responses from 106 human participants. While some models, especially Claude 3.7 Sonnet, approximated human averages and lexical preferences, none reproduced the scope of human variability. LLM outputs were consistently less diverse, with newer models and thinking-enabled modes often reducing rather than increasing variability. Network analysis further revealed fundamental differences in retrieval structure between humans and the most human-like model. Ensemble simulations combining outputs from diverse models also failed to recover human-level diversity, likely due to high vocabulary overlap across models. These results highlight key limitations in using LLMs to simulate human cognition and behavior.
\\ \newline 
}

\begin{document}

\maketitleabstract

\section{Introduction}

Large language models (LLMs) have rapidly advanced in recent years, achieving impressive performance across a wide range of natural language tasks. As a result, researchers have become increasingly interested in using LLMs as experimental tools in cognitive and behavioral science. Some even propose that LLMs could replace human participants in certain studies, offering scalable and efficient alternatives for simulating human behavior \citep{dillion2023can}.

This idea has intuitive appeal: LLMs can generate fluent responses on demand, and arguably encode the ``wisdom of the crowd'' from their massive training data \citep{trott2024large}. For instance, \citet{hansen2022semantic} found that GPT-3 could generate semantic features for concepts that not only mirrored the distribution of human-generated features but also matched human norms in their ability to predict similarity, relatedness, and category membership. On the other hand, a growing body of work has pointed out that LLMs may lack a crucial feature of human language and cognition: variability. \citet{cuskley2024limitations} argue that LLMs are limited by their training on written language, which captures only a narrow slice of human communicative behavior. \citet{zanotto2024human} offer empirical support, showing that LLM-generated texts exhibit substantially lower linguistic diversity than human-written texts across a range of features, including syntax, vocabulary, and style.

This lack of variability is also evident when examining how LLMs perform in tasks designed to probe semantic or associative structures. In a semantic fluency task where participants named as many animals as possible, \citet{wang-etal-2025-fluency} found that LLM-generated semantic networks were structurally different from those of humans. Compared to human networks, those of LLMs exhibited weaker local associations, poorer global integration, and greater rigidity in semantic organization. Similarly, \citet{haim2025cognitive} examined word association networks in a STEM-related mindset task and found that GPT-3.5 produced networks that were notably sparser and less interconnected than those generated by human participants.

It is worth noting that \citet{wang-etal-2025-fluency} recognized this limitation and attempted to address it by prompting LLMs to role-play 30 different occupations, effectively simulating multiple distinct participants. However, the resulting semantic networks still failed to match the flexibility and associative structure of human data. Crucially, real human participants, even those drawn from relatively homogeneous groups such as college students, consistently exhibit meaningful individual differences. These differences may reflect not only variation in the amount, type, and organization of semantic knowledge, but also differences in executive function, and strategy use. Of course, because LLMs are optimized to encode and reproduce stable semantic relationships, consistency across prompts may reflect a design feature. From this perspective, the internal coherence and semantic regularity that make LLMs so effective at language modeling may also constrain their ability to simulate the variability that characterizes human cognition.

If semantic-based tasks limit LLMs' ability to simulate human behavioral variability, it raises the question: how would they perform on a task that relies less on meaning? One such task is phonemic (or letter) fluency, where individuals are asked to produce as many words as possible that begin with a particular letter, such as \textit{F}, within a fixed time limit. Unlike semantic fluency, this task requires an effortful search through the mental lexicon based on orthographic or phonological form--an unnatural retrieval strategy in everyday language use. Prior analyses have shown that phonemic fluency performance is influenced by both lexical properties, such as word frequency, age of acquisition, and phonetic similarity, and cognitive factors like switching patterns \citep{cho-etal-2021-automated}.

The goal of the present study is to explore whether LLMs can simulate inter-participant variability in phonemic fluency, or whether their outputs remain rigid, perhaps driven by frequency-based or alphabetic heuristics. Unlike previous work that introduced arbitrary personas such as fictional occupations \citep{wang-etal-2025-fluency}, we ground our simulations in real participant metadata. Extensive research in psychology has consistently shown that age and education are among the most reliable predictors of verbal fluency performance \citep{tombaugh-etal-1999-normative, gladsjo-etal-1999-norms, villalobos-etal-2023-systematic}. By providing LLMs with demographic and performance information drawn directly from real human participants, the simulation can be more meaningfully compared against actual human data.

Furthermore, previous studies comparing LLM and human cognitive and language performance have typically relied on one or a small number of models. The present study offers a substantially more comprehensive evaluation, testing $34$ distinct models across $45$ configurations from major closed-source and open-source providers. This broad coverage allows us to identify patterns that hold across model families and architectures, rather than drawing conclusions from a single system. 

It also enables us to address a fundamental concern about LLM-based simulation: using the same model to simulate multiple participants means those simulated individuals inherently share the same foundational knowledge, making it difficult to generate authentically varied outputs regardless of prompt manipulation \citep{wang2025canllmsimulations}. To overcome this limitation, we investigate whether combining outputs from diverse models through ensemble sampling---where each simulated participant's response is drawn from a randomly selected model---can better approximate human-level diversity, leveraging architectural and training differences across models to collectively recover variability that no single model captures alone.

In preview, our findings suggest a clear answer: LLMs are highly capable of producing fluent and correct responses in the phonemic fluency task, but none of them---individually or in ensemble---match the full scope of human behavioral variability. The remainder of the paper is organized as follows. Section~\ref{sec:method} describes the human \textit{F} fluency dataset and the LLM simulation procedure. Section~\ref{sec:participant} evaluates participant-level response counts and lexical diversity. Section~\ref{sec:item} presents item-level analyses of production frequency distributions and linguistic predictors. Section~\ref{sec:network} compares word co-occurrence network structures. Section~\ref{sec:ensemble} examines ensemble-based diversity. Section~\ref{sec:discussion} concludes with a discussion of implications and future directions.

\section{Method}\label{sec:method}

\subsection{Human data source}

We used phonemic fluency data collected by \citet{qiu-johns-2021-vf}\footnote{The verbal fluency data are publicly available on the Open Science Framework under the Creative Commons Attribution 4.0 International License (CC BY 4.0).}, who investigated noun- and verb-based semantic fluency across two experiments (see also \citealp{qiu-etal-2021-vf}). In both experiments, participants first completed a phonemic fluency task using the letter \textit{F} as a familiarization trial before proceeding to semantic tasks. A total of 106 native English speakers (age: $M = 35.59, SD = 10.04$; education: $M = 14.92$ years, $SD = 2.01$) were recruited via Amazon Mechanical Turk and completed the tasks online. Participants were instructed to produce as many words as possible beginning with the letter \textit{F} within one minute, excluding proper nouns, numbers, and same-lemma variants. Audio recordings were collected for all participants.

Following the original study's procedure, we used the Google Speech-to-Text API\footnote{\url{https://cloud.google.com/speech-to-text}} to transcribe the recordings, followed by manual review to exclude errors such as repetitions, out-of-category responses, same-lemma variants (e.g., \textit{fish}, \textit{fishing}, \textit{fished}), and counting behavior. After corrections, participants produced an average of 16.89 words ($SD = 4.84$).

\subsection{LLM simulation procedure}

We evaluated a diverse set of closed-source and open-source LLMs. For closed-source models, we used the official APIs from OpenAI\footnote{\url{https://openai.com/api/}}, Anthropic\footnote{\url{https://www.anthropic.com/api}}, Google\footnote{\url{https://ai.google.dev/gemini-api/docs}}, and xAI\footnote{\url{https://x.ai/api}}. For open-source models, we accessed models from Meta, Moonshot, Alibaba, Zhipu AI, DeepSeek, and Mistral via the Together.ai API\footnote{\url{https://www.together.ai}}.

All models received the same full prompt, which included the participant's age, education level, and number of correct responses, along with task instructions identical to those given to human participants (see Appendix~\ref{app:prompt} for the complete prompt). The prompt instructed models to role-play as the corresponding human participant and generate words beginning with \textit{F} under a simulated one-minute constraint, outputting one word per line with no additional text. Each model was prompted independently 106 times, once per human participant, with each prompt constituting a separate API call with no memory of previous calls.

In total, we evaluated 34 distinct models. Given that many recent models support hybrid thinking or reasoning modes, we additionally tested these models with and without extended thinking enabled, and at different reasoning effort levels where applicable, resulting in 45 configurations.

\section{Participant-Level Analysis}\label{sec:participant}

\subsection{Number of responses}

Human participants produced an average of 16.89 correct responses within the one-minute time constraint ($SD = 4.84$). To evaluate whether LLMs adhered to the performance constraints specified in the prompt, we computed the Mean Absolute Error (MAE) between LLM-generated and human response counts for each configuration. Configurations with $\text{MAE} \leq 1.69$ (i.e., within 10\% of the human mean) were considered to have successfully simulated human-like production rates.

Of the 45 configurations tested, 33 met this criterion, indicating that the majority of models generally adhered to the instructed constraints. The remaining 12 configurations significantly overproduced relative to human participants. Among these, OpenAI's GPT-4 Turbo ($\text{MAE} = 57.17$) and o3 ($\text{MAE} = 35.39$) showed the most extreme overproduction, generating outputs several times longer than human responses. Surprisingly, two of Anthropic's latest models, Claude 4.6 Sonnet (Thinking) ($\text{MAE} = 3.19$) and Claude 4.6 Opus ($\text{MAE} = 3.87$), also failed to adhere to the performance constraints despite the explicit inclusion of the number of correct responses in the prompt.

Among the 33 configurations with $\text{MAE} \leq 1.69$, we further examined whether any individual simulations deviated substantially from the corresponding human participant's response count. Configurations containing any simulation where $|\text{LLM}-\text{Human}| > 5$ were flagged as having outliers. Of the 33 configurations, 21 produced no outliers across all 106 simulations and were retained for subsequent analyses. Details of included and excluded configurations are provided in Appendix~\ref{app:model_selection}.

\subsection{Variability of responses}

To evaluate the variability of responses across human participants and LLM-simulated participants, we computed the Type-to-Token Ratio (TTR) and the Idiosyncratic Type-to-Total Type Ratio (ITTTR). TTR captures lexical diversity by dividing the number of unique words (types) by the total number of responses (tokens). A higher TTR indicates a wider range of vocabulary produced. ITTTR measures how many of those types were idiosyncratic (i.e., produced by only one participant) relative to the total number of types. This provides insight into how consistent or individualized the generated words are across participants within the group \citep{castro-etal-2021-category}.

As shown in Table~\ref{tab:types}, human participants produced the highest number of unique types ($476$) and idiosyncratic types ($201$), resulting in a TTR of $0.27$ and an ITTTR of $0.42$. No LLM approached this level of variability. The most diverse LLM output came from Claude 3.7 Sonnet ($226$ types, TTR $= 0.13$, ITTTR $= 0.32$), followed by o3-mini ($193$ types, TTR $= 0.11$, ITTTR $= 0.30$). Even these best-performing models produced fewer than half the unique types observed in human data.

Several patterns emerged across model families. For Anthropic models, non-thinking configurations consistently produced more diverse outputs than their thinking-enabled counterparts (e.g., Claude 3.7 Sonnet: $226$ vs.\ $137$ types; Claude 4.5 Sonnet: $110$ vs.\ $77$ types; Claude 4 Sonnet: $74$ vs.\ $73$ types). Additionally, the earlier Claude 3.7 Sonnet substantially outperformed more recent Anthropic models in lexical diversity, suggesting that newer models may prioritize consistency over variability.

For OpenAI models, reasoning capacity appeared to benefit diversity. GPT-5.2 showed a clear gradient, with higher reasoning effort producing more unique types (High: $158$, Medium: $149$, Low: $139$). The most diverse OpenAI model, o3-mini ($193$ types), is itself a reasoning-focused model. Despite these differences across models, all LLM configurations remained well below human-level variability in both TTR and ITTTR, highlighting a persistent tendency toward output rigidity across architectures and providers.

\begin{table}[htbp]
\centering
\caption{Type-to-token ratios (TTR) and idiosyncratic type-to-total type ratios (ITTTR) across human participants and LLM-simulated participants}
\label{tab:types}
\renewcommand{\arraystretch}{1.3}
\resizebox{\columnwidth}{!}{%
\begin{tabular}{lrrrrr}
\hline
\textbf{Model} & \textbf{Types} & \textbf{Idio. Types} & \textbf{Tokens} & \textbf{TTR} & \textbf{ITTTR}\\
\hline
\textbf{Human} & \textbf{476} & \textbf{201} & \textbf{1790} & \textbf{0.27} & \textbf{0.42}\\
\hdashline
\multicolumn{6}{l}{\textit{OpenAI}} \\
o3-mini & 193 & 58	& 1790	& 0.11 & 0.30 \\
GPT-5.2 (High Reasoning) & 158	& 40 & 1790	& 0.09 &	0.25 \\
GPT-5.2 (Medium Reasoning) & 149	& 38	& 1790	& 0.08 &	0.26 \\
GPT-3.5 Turbo & 148	& 48	& 1783	& 0.08	& 0.32 \\
GPT-5.2 (Low Reasoning) & 139	& 35	& 1790	& 0.08 &	0.25 \\
GPT-5 & 130	& 33 & 1790	& 0.07 & 0.25 \\
GPT-4o & 128 & 38	& 1773	& 0.07 & 0.30 \\
GPT-5 mini & 95	& 18	& 1790	& 0.05	& 0.19 \\
GPT-4.1 & 73 & 20	& 1790	& 0.04 & 0.27 \\
\hdashline
\multicolumn{6}{l}{\textit{Anthropic}} \\
Claude 3.7 Sonnet & 226	& 73 & 1781	& 0.13	& 0.32 \\
Claude 3.7 Sonnet (Thinking) & 137	& 43 & 1788	& 0.08	& 0.31 \\
Claude 4.5 Sonnet & 110	& 32 & 1784	& 0.06	& 0.29 \\
Claude 4.5 Sonnet (Thinking) & 77	& 18	& 1790	& 0.04 &	0.23 \\
Claude 4 Sonnet & 74 & 12	& 1795	& 0.04 &	0.16 \\
Claude 4 Sonnet (Thinking) & 73	& 18	& 1789	& 0.04 &	0.25 \\
\hdashline
\multicolumn{6}{l}{\textit{Google}} \\
Gemini 2.5 Flash & 149	& 51	& 1794	& 0.08	& 0.34 \\
Gemini 2.5 Pro & 83	& 20 & 1790	& 0.05 & 0.24 \\
Gemini 3 Pro & 73	& 20	& 1790	& 0.04 &	0.27 \\
\hdashline
\multicolumn{6}{l}{\textit{xAI}} \\
Grok-4.1-fast (Reasoning) & 62	& 8	 & 1787	 & 0.03 &	0.13 \\
\hdashline
\multicolumn{6}{l}{\textit{Open-Source Models}} \\
Kimi K2  & 96	& 25	& 1787	& 0.05	& 0.26 \\
Qwen3-2507 & 93	& 22 & 1781	& 0.05 &	0.24 \\
\hline
\end{tabular}}
\end{table}

\section{Item-Level Analysis}\label{sec:item}

\subsection{Distribution of production frequency}

Previous studies have observed that word production in verbal fluency tasks typically follows Zipf's law, a type of power-law relationship where word frequency is inversely proportional to its rank, expressed as $f(r) \propto 1/r^{\alpha}$, where $f(r)$ denotes the frequency of a word at rank $r$ and $\alpha$ is the scaling exponent \citep[e.g.,][]{taler-etal-2020-large}. This distribution is characterized by a small number of words that are produced frequently across participants, followed by a ``long tail'' of words that are produced infrequently, often appearing only once or twice in the entire dataset. 

\begin{figure*}[htbp] \centering \includegraphics[width=\textwidth]{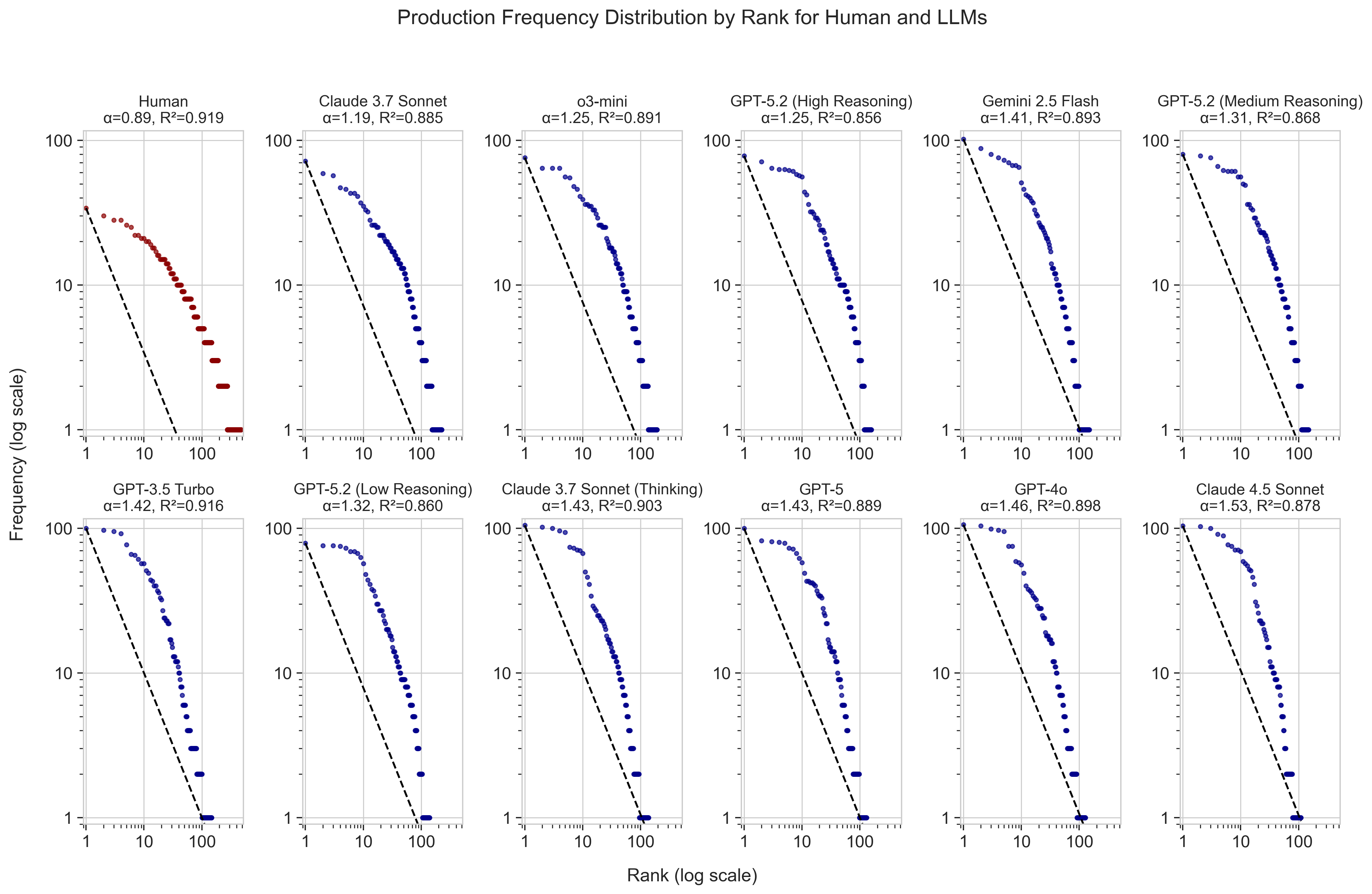} \caption{Production frequency distribution by rank for human participants (red) and $11$ LLM configurations (blue) with at least $100$ unique types. Both axes are log-scaled. Dashed lines represent fitted power-law models, with $\alpha$ and $R^2$ shown for each panel.} \label{fig:zipf} \end{figure*}

To estimate $\alpha$, we performed linear regression on the log-transformed rank and frequency values; the slope of the fitted line provides the $\alpha$ estimate, and $R^2$ indicates goodness of fit. A larger $\alpha$ reflects a steeper decline, indicating that top-ranked words dominate more heavily with less diversity in the long tail.
Only model configurations that produced at least $100$ unique types were included in this analysis, resulting in $11$ configurations alongside the human data (Figure~\ref{fig:zipf}).

Human participants' word production closely followed the expected Zipfian pattern, with $\alpha = 0.89$ and $R^2 = 0.92$. The most frequently produced word was ``fun'', mentioned $34$ times, followed by ``fire'' ($30$ times) and ``fan'' ($28$ times). In contrast, LLMs demonstrated a notably different pattern. While all models showed adequate fit to a power-law distribution ($R^2$ ranging from $0.86$ to $0.92$), their consistently higher $\alpha$ coefficients ($1.19$--$1.53$) reflect steeper declines in frequency, with top-ranked words dominating the distribution more heavily than in human data. Among the LLMs, Claude 3.7 Sonnet exhibited the closest $\alpha$ to the human distribution ($\alpha = 1.19$, $R^2 = 0.89$), though still notably steeper than the human pattern.

\subsection{Linguistic variables and production frequency}

Previous research has shown that verbal fluency performance is influenced by multiple linguistic variables, including word frequency, familiarity, word duration, and age of acquisition \citep{cho-etal-2021-automated, taler-etal-2020-large}. In this study, we examined the relationship between production frequency and core linguistic variables obtained from the English Lexicon Project \citep{balota2007english}, including word length (number of letters), word frequency (log frequency per million words from SUBTLEX corpus), orthographic neighborhood size (number of words that differ by one letter), phonological neighborhood size (number of words that differ by one phoneme), semantic neighborhood size (number of semantically related words), and age of acquisition (estimated age when a word is typically learned). As with the Zipf analysis, only model configurations with at least $100$ unique types were included, resulting in $11$ configurations alongside human data. We first conducted simple correlation analyses between these variables and production frequency.

As shown in Table~\ref{tab:correlation}, all linguistic variables except for semantic neighborhood size were significantly correlated with production frequency in human data. Word frequency showed the strongest positive correlation ($r = 0.55$), followed by age of acquisition ($r = -0.52$) and orthographic neighborhood size ($r = 0.43$). Among the LLMs, Claude 3.7 Sonnet most closely approximated the human pattern, with highly similar correlation coefficients across all variables. o3-mini followed a similar trend but with generally weaker correlations, particularly for orthographic neighborhood size. The remaining models showed progressively weaker alignment, with many losing significance for neighborhood variables while retaining word frequency and age of acquisition as reliable predictors. Semantic neighborhood size remained non-significant across all models, consistent with the phonemic nature of the task.

\begin{table*}[htbp]
    \centering
    \caption{Correlations between linguistic variables and production frequency across models}
    \label{tab:correlation}
    \sisetup{table-number-alignment = center, table-figures-integer=1, table-figures-decimal=2}
     \resizebox{0.95\textwidth}{!}{%
    \begin{tabular}{l S[table-format=-1.2] S[table-format=-1.2] S[table-format=-1.2] S[table-format=-1.2] S[table-format=-1.2] S[table-format=-1.2]}
        \hline
        \textbf{Model} & \textbf{Length} & \textbf{WF} & \textbf{Ortho\_N } & \textbf{Phono\_N} & \textbf{Sem\_N} & \textbf{AoA}\\
        \hline
        Human      & -0.39$^{***}$  & 0.55$^{***}$  &  0.43$^{***}$   & 0.38$^{**}$  & 0.01 & -0.52$^{***}$ \\
        \hdashline
        Claude 3.7 Sonnet &  -0.34$^{***}$  &  0.54$^{***}$  & 0.33$^{***}$   &  0.31$^{***}$ & -0.02  & -0.49$^{***}$ \\
        o3-mini    &  -0.29$^{***}$  &  0.40$^{***}$  & 0.24$^{**}$   &  0.36$^{***}$ & -0.004 & -0.36$^{***}$   \\
        GPT-5.2 (High Reasoning) & -0.05 & 0.31$^{***}$ & -0.06 & -0.01 & -0.03 & -0.37$^{***}$ \\
        Gemini 2.5 Flash & -0.18$^{*}$ & 0.43$^{***}$ & 0.17$^{*}$ & 0.16$^{*}$ & -0.03 & -0.51$^{***}$ \\
        GPT-5.2 (Medium Reasoning) & -0.07 & 0.31$^{***}$ & -0.05 & -0.04 & -0.04 & -0.39$^{***}$ \\
        GPT-3.5 Turbo & -0.16 & 0.30$^{***}$ & 0.16 & 0.14 & 0.08 & -0.44$^{***}$ \\
        GPT-5.2 (Low Reasoning) & 0.003 & 0.25$^{***}$ & -0.10 & -0.12 & -0.04 & -0.32$^{***}$ \\
        Claude 3.7 Sonnet (Thinking) &  -0.23$^{**}$  &  0.43$^{***}$  & 0.23$^{**}$   &  0.20$^{*}$ & -0.10  & -0.43$^{***}$ \\
        GPT-5 & -0.09 & 0.30$^{***}$ & -0.01 & -0.07 & 0.04 & -0.48$^{***}$ \\
        GPT-4o & -0.04 & 0.03 & -0.05 & -0.05 & 0.10 & -0.27$^{**}$ \\
        Claude 4.5 Sonnet &  -0.15  &  0.35$^{***}$  & 0.12   &  0.09 & -0.05  & -0.50$^{***}$ \\
        \hline
        
        \multicolumn{7}{l}{\footnotesize \parbox{\textwidth}{
        \vspace{0.5em}
        Note. Length = word length; WF = log word frequency (SUBTLEX); Ortho\_N = orthographic neighborhood size; Phono\_N = phonological neighborhood size; Sem\_N = semantic neighborhood size; AoA = age of acquisition.\\ $^{***}p < .001$; $^{**}p < .01$; $^{*}p < .05$}} \\
    \end{tabular}
    }
\end{table*}

To formally establish the factors that influence production frequency, we conducted stepwise regression analyses with all linguistic variables. Full statistical details are provided in Appendix~\ref{app:regression}, but a consistent pattern emerged: linguistic variables explained the most variance in production frequency for human responses (adjusted $R^2 = .396$), followed closely by Claude 3.7 Sonnet (adjusted $R^2 = .356$), with substantially less variance explained for o3-mini (adjusted $R^2 = .179$). Claude 3.7 Sonnet most closely matched the human pattern in both variance explained and predictors retained, with word frequency and age of acquisition emerging as significant in both. Orthographic neighborhood size was uniquely important for humans, while word length played a stronger role in Claude 3.7 Sonnet and o3-mini.

\section{Network Analysis}\label{sec:network}

Building on our earlier findings, we focused the network analysis exclusively on comparing human and Claude 3.7 Sonnet outputs. Claude consistently exhibited the most human-like patterns--response count, variability, production frequency distribution, and linguistic predictors of word choice.

\subsection{Network construction approach}

To better understand the organizational structure of word production in letter \textit{F} fluency, we constructed correlation-based networks---a common approach in cognitive science for examining the relationships between concepts \citep{borodkin-etal-2016-pumpkin,li-qiu-2025-semantic,siew-guru-2023-investigating}. This approach is based on the premise that word co-occurrence patterns in fluency tasks reflect underlying associative relationships: the tendency to produce word $b$ given that word $a$ was also produced indicates a shared associative link between those items. By aggregating these co-occurrence patterns across participants, it is possible to estimate a network structure where nodes represent words and edges capture the strength of association \citep{borodkin-etal-2016-pumpkin}.

Given the different number of nodes in the full networks ($476$ words for humans, $226$ for Claude), direct comparison of network metrics may be confounded by size differences \citep{borodkin-etal-2016-pumpkin}. Therefore, we selected $197$ words that were produced by both humans and Claude to ensure a fairer comparison.

The network construction involved three steps: (1) creating binary participant-by-word matrices indicating which words each participant produced, (2) computing word-word cosine similarity matrices based on co-occurrence patterns across participants, and (3) applying the Triangulated Maximally Filtered Graph (TMFG; \citealp{massara2017network}) algorithm to retain only the strongest associations while preserving network connectivity.

\subsection{Structural comparison of human and Claude networks}

We compared the structural properties of networks constructed from human and Claude 3.7 Sonnet outputs using two standard global metrics: clustering coefficient (CC) and average shortest path length (ASPL). CC quantifies the degree to which nodes in a network tend to cluster together (i.e., the likelihood that two neighbors of a node are also connected), while ASPL reflects the average number of steps required to connect any two nodes, indexing the global efficiency of the network \citep{siew2019cognitive}.

Both networks were constructed from the same set of 197 shared words to control for network size. The human network showed a higher CC ($0.42$) and longer ASPL ($5.32$) compared to Claude ($\text{CC} = 0.37$, $\text{ASPL} = 4.40$). To assess whether the observed structures differed from random expectations, we simulated 1,000 Erd{\H{o}}s-R{\'e}nyi random networks \cite{erdHos1960evolution} matched in size and density. One-sample $z$-tests confirmed that both human and Claude networks significantly deviated from randomness, exhibiting substantially higher CCs and slightly higher ASPLs than random networks (human: $z = -2460.1$ for CC, $z = -3765.1$ for ASPL; Claude: $z = -2105.2$ for CC, $z = -2410.6$ for ASPL; all $p < .001$).

\begin{figure}
\centering
    \includegraphics[width=\columnwidth]{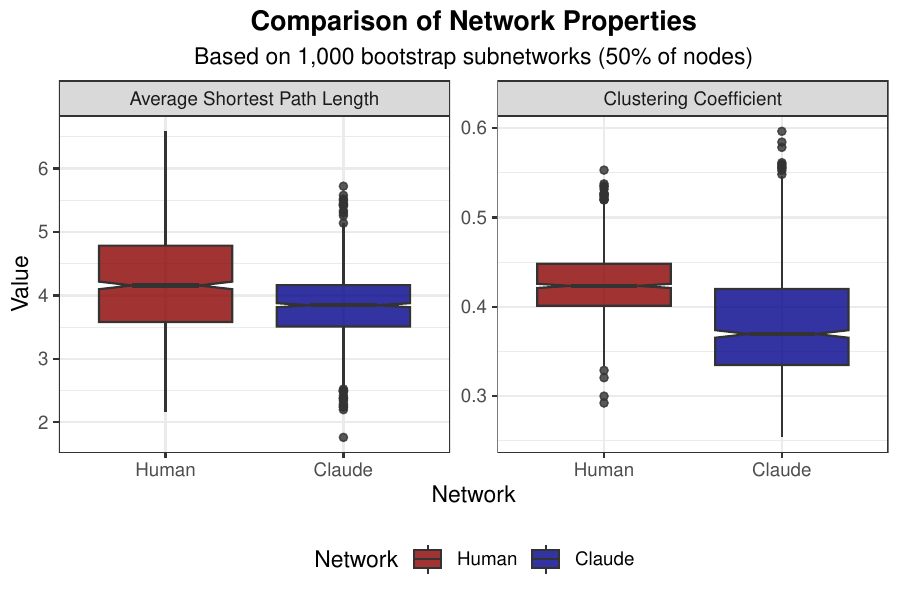}  % Adjust width if needed
    \caption{Clustering coefficient and average shortest path length for human and Claude networks based on 1,000 bootstrap subnetworks (50\% nodes)}
    \label{fig:network}
\end{figure}

To statistically compare the two networks, we conducted a subnetwork bootstrap procedure: 1,000 partial networks were generated for each system by randomly sampling 50\% of nodes and recalculating network metrics \citep{borodkin-etal-2016-pumpkin,qiu-etal-2021-vf}. As shown in Figure~\ref{fig:network}, human subnetworks had significantly higher CCs ($t = 19.71$, $p < .001$) and longer ASPLs ($t = 11.48$, $p < .001$) than Claude subnetworks. These differences indicate that human responses are organized into tighter local clusters with weaker global integration, whereas Claude responses form a more evenly connected network with greater global efficiency.

To complement the network analysis, we conducted a representational similarity analysis (RSA; \citealp{nili2014toolbox}) comparing the original pairwise similarity matrices that underlie network construction. The Spearman correlation between the lower triangular portions of the human and Claude matrices was modest ($\rho = .22$, $p < .001$), providing additional evidence that Claude organizes or retrieves words differently than humans. Taken together with the network metric results, this suggests that while Claude can produce human-like words and respond to similar lexical variables, the global structure and retrieval dynamics it exhibits remain meaningfully distinct from human performance.

\section{LLM Ensemble Analysis}\label{sec:ensemble}

The preceding analyses demonstrate that no single LLM configuration captures the full scope of human behavioral variability. A natural question is whether combining outputs from multiple models---each potentially drawing on different architectures, training data, and alignment strategies---could better approximate the distributional diversity observed in human responses. While recent ensemble methods in NLP have primarily focused on improving task accuracy through techniques such as majority voting \citep{li2024more}, our goal is the opposite: to explore whether ensembling can simulate the distributional variability characteristic of human behavior.

\subsection{Ensemble construction}

To construct ensemble outputs, we used a participant-level random sampling procedure. For each of the $106$ human participants, we assembled a pool of eligible LLM outputs from all $33$ configurations with $\text{MAE} \leq 1.69$, excluding any outlier simulations for that participant (see Appendix~\ref{app:model_selection}). This resulted in $29$ to $33$ available LLM outputs per participant, depending on how many configurations produced outliers for that individual.

For each ensemble simulation, we independently and randomly selected one LLM output per participant from their available pool, producing $106$ simulated participants from a mixture of models. This procedure was repeated $1{,}000$ times, yielding $1{,}000$ ensemble samples.

\subsection{Ensemble results}

\begin{figure*}[htbp] \centering \includegraphics[width=\textwidth]{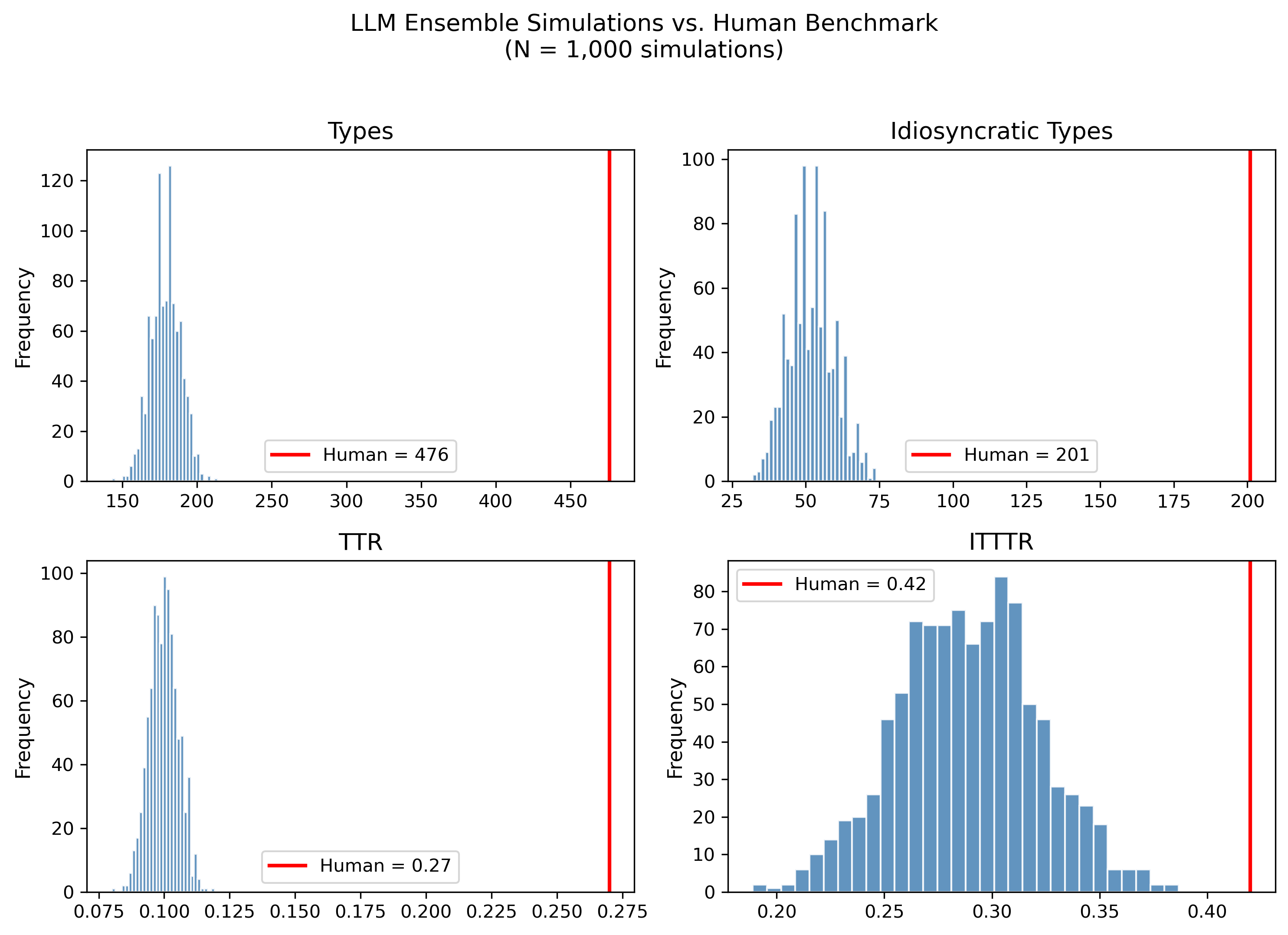} \caption{Distributions of variability metrics across $1{,}000$ LLM ensemble simulations. Red vertical lines indicate human values. All ensemble distributions fall substantially below human benchmarks.} \label{fig:ensemble} \end{figure*}

Figure~\ref{fig:ensemble} shows the distributions of variability metrics across the $1{,}000$ ensemble simulations, with human values indicated by the red vertical lines. Across all metrics, the ensemble distributions fell far short of human-level variability. The mean number of unique types across simulations was $179.13$ ($SD = 9.82$, range: $143$--$214$), compared to $476$ for human participants. Similarly, mean TTR was $0.100$ ($SD = 0.005$, range: $0.080$--$0.119$), well below the human TTR of $0.27$. The same pattern held for idiosyncratic types and ITTTR.

Notably, the ensemble simulations did not substantially exceed the diversity of the best-performing individual model. Claude 3.7 Sonnet alone produced $226$ unique types with a TTR of $0.13$---values that fall within or near the upper tail of the ensemble distribution for types but above the ensemble range for TTR. 

The Zipf analysis of ensemble outputs reinforced this conclusion. Across $1{,}000$ simulations, the mean $\alpha$ coefficient was $1.29$ ($SD = 0.03$, range: $1.18$--$1.37$), substantially steeper than the human value of $\alpha = 0.89$ and comparable to individual model values (e.g., Claude 3.7 Sonnet: $\alpha = 1.19$; o3-mini: $\alpha = 1.25$).

\subsection{Vocabulary overlap across models}

To understand why ensembling failed to increase diversity, we computed pairwise vocabulary overlap coefficients between all $21$ models retained for participant-level analyses (see Appendix~\ref{app:overlap} for the full pairwise matrix). The overlap coefficient, defined as the size of the intersection divided by the size of the smaller vocabulary, was remarkably high across the board (mean $= 0.74$, $SD = 0.11$). 

Within Anthropic models, overlap was especially pronounced (mean $= 0.90$), with some pairs reaching $0.99$, indicating that switching between model versions barely changes the word pool. OpenAI models showed somewhat more internal diversity (mean $= 0.72$), though this was largely driven by GPT-3.5 Turbo, an older architecture with the lowest overlap with other OpenAI models. 

Even across providers, roughly three-quarters of any model's vocabulary appeared in every other model. This pervasive lexical overlap explains why combining outputs from diverse models yields minimum gains in diversity: the models are drawing from a largely shared word pool, and ensembling amplifies high-frequency core vocabulary rather than expanding into the long tail of rare or idiosyncratic responses.

\section{Discussion and Conclusion}\label{sec:discussion}

This study evaluated whether LLMs can simulate the behavioral variability observed in human phonemic fluency performance. Across $34$ models and $45$ configurations, no model---individually or in ensemble---captured the full scope of human variability, despite many producing plausible average response counts and showing sensitivity to similar linguistic features.

A key contribution of this study is its comprehensive coverage of models. Our results reveal that LLMs do not perform equally well on this task, and importantly, newer models are not necessarily better. Within Anthropic, the earlier Claude 3.7 Sonnet substantially outperformed more recent models in lexical diversity, and enabling thinking mode consistently reduced rather than increased variability. These findings suggest that researchers conducting LLM-based simulations of human behavior should evaluate multiple models across different providers rather than defaulting to the latest release, as model selection can substantially influence the conclusions drawn.

Beyond differences in variability, our analyses also revealed divergent retrieval mechanisms. At the item level, orthographic neighborhood size played a unique role for human participants, whereas LLMs showed greater sensitivity to word length, suggesting that humans rely more on form-based associations during phonemic retrieval while LLMs prioritize more surface-level features. The network analyses reinforced this divergence: human responses exhibited stronger local clustering but weaker global integration, whereas Claude 3.7 Sonnet optimized for network-wide efficiency. Together, these structural mismatches highlight a core limitation---LLMs can produce plausible words but fail to approximate the associative dynamics of human memory search.

A fundamental concern motivating our ensemble analysis was that using the same model to simulate multiple participants means those simulated individuals share the same foundational knowledge, making it difficult to generate authentically varied outputs \citep{wang2025canllmsimulations}. Our ensemble approach addressed this by drawing each simulated participant's response from a randomly selected model. Yet even this strategy failed: the pairwise vocabulary overlap analysis revealed that models across providers draw from a largely shared word pool, with within-family overlap reaching $0.90$ for Anthropic models. This convergence likely reflects the fact that current LLMs are fundamentally trained on similar internet-scale data using transformer-based architectures, leading to shared biases that amplify high-frequency, prototypical responses at the expense of the rare and idiosyncratic words that characterize human individuality. This pattern raises broader concerns about representational bias: if LLMs systematically suppress low-frequency responses, they risk reinforcing dominant linguistic patterns while marginalizing the diversity inherent in real human populations \citep{guo-etal-2024-bias}.

Among all models tested, Claude 3.7 Sonnet consistently emerged as the most human-like, and performed far better than other configurations. Why this particular model outperforms others, including more recent models from the same provider, remains an open question. Future work could focus on extending this model's capabilities through more targeted persona modeling, potentially incorporating linguistic and cognitive profiles (e.g., selective reading exposure; \citealp{qiu-etal-2024-apt}) to better simulate participant-level differences. More broadly, rather than substituting LLM-simulated individuals for real human participants, a more productive approach may be to treat LLMs as a baseline system--one that captures central tendencies in linguistic knowledge--against which the uniqueness and flexibility of real human behavior can be measured.

In sum, while LLMs continue to impress in terms of fluency and surface-level plausibility, our findings highlight the challenges of using them as substitutes for human participants in behavioral research--especially when individual variability and cognitive process modeling are central. Phonemic fluency reveals these limitations clearly, offering a strong case for developing new strategies if LLMs are to be used meaningfully in the simulation of human behavior.

\section*{Limitations}

This study focuses on a single verbal fluency task (letter \textit{F}) in English, limiting generalizability across task types, letters, and languages. While phonemic fluency provides a stringent test of form-based retrieval, different letters may elicit distinct lexical or structural patterns. Similarly, results may differ in languages with different orthographic or phonological systems. Our analysis also centers on a specific set of models and parameter settings; other prompting techniques (e.g., few-shot examples) or fine-tuned models may yield different patterns. Additionally, while we grounded our simulations in real participant metadata, age and education are not the only sources of individual variation in verbal fluency. Factors such as profession, domain expertise, and accumulated life experience also contribute to behavioral differences, and incorporating these richer individual-level traits may be necessary to drive more human-like variability. Future work should expand to multilingual and cross-task comparisons and evaluate whether alternative sampling or persona modeling strategies better capture human behavioral diversity.

\section*{Ethical Considerations}
The human data used in this study are drawn from \citet{qiu-johns-2021-vf}, a publicly available dataset under a CC-BY 4.0 license. The dataset includes anonymized verbal fluency responses alongside participant age and education information, with all personal identifiers removed. According to the original publication, all data collection procedures received institutional ethics approval, and participants provided informed consent. In our study, only de-identified demographic and performance information was used to construct LLM prompts. At no stage do we include real human response content in the prompts, nor do we attempt to reconstruct the linguistic patterns of specific individuals.

Our findings also raise ethical concerns regarding the use of LLMs to simulate human behavioral variability. LLMs exhibit output rigidity, favoring high-frequency, prototypical responses and failing to capture natural human diversity. If used as substitutes for human participants without acknowledging this limitation, LLM-based simulations risk systematically underrepresenting the variability present in real populations, particularly marginalizing rare or idiosyncratic responses. We therefore caution against treating LLM-simulated individuals as direct substitutes for real human participants, especially in contexts where diversity or individual identity are central to the research question.

\section*{Bibliographical References}\label{sec:reference}
\bibliographystyle{lrec2026-natbib}
\bibliography{ref,ref2}
%\onecolumn
% \newpage
\appendix
\setcounter{table}{0}
\renewcommand{\thetable}{A\arabic{table}}

\setcounter{figure}{0}
\renewcommand{\thefigure}{A\arabic{figure}}

\section{Prompt Design}\label{app:prompt}

Figure~\ref{fig:prompt-code} shows the complete prompt used to simulate the \textit{F} fluency task. The prompt included three pieces of participant-level information: age, education level, and number of correct responses.

Age and education were included because they are among the most reliable predictors of verbal fluency performance, consistently accounting for significant variance in both phonemic and semantic fluency across languages and populations \citep{tombaugh-etal-1999-normative, gladsjo-etal-1999-norms, villalobos-etal-2023-systematic}. These variables were drawn directly from the human dataset rather than fabricated, ensuring ecological validity in the simulation.

In terms of the inclusion of the number of correct responses, \citet{wang-etal-2025-fluency} found that LLMs produced comparable numbers of animal names to human participants when prompted with only a one-minute time constraint and no performance information, suggesting that LLMs can approximate typical human production rates for semantically driven tasks. However, in our pilot testing of the phonemic fluency task, we found that prompting models with only age and education information led to significant overproduction. This likely reflects the greater difficulty of phonemic fluency for humans, who must perform an effortful, non-semantic search through the mental lexicon---a constraint that does not apply to LLMs. To ensure realistic response counts and enable meaningful comparisons at the item and network levels, we therefore also provided the number of correct responses as an additional constraint in the prompt.

\begin{figure*}[htbp]
    \centering
    \begin{lstlisting}[
        language=Python, 
        basicstyle=\footnotesize\ttfamily, 
        breaklines=true, 
        showstringspaces=false,
        backgroundcolor=\color{lightgray!20},
        frame=single,               % Adds a single-line border
        framexleftmargin=10pt,      % Adds some margin inside the left border
        framexrightmargin=10pt,     % Adds some margin inside the right border
        xleftmargin=10pt,           % Adds some margin outside the left border
        xrightmargin=10pt           % Adds some margin outside the right border
    ]
prompt = (
    "In a verbal fluency task, you will be asked to say as many words as you can think of that conform to a specified criterion within 1 minute. "
    "Please DO NOT say words that are proper nouns (like Bob or Boston), numbers, or the same words with different endings (for example, love -> loves, lover, loving). Please DO NOT say phrases or sentences. Please DO NOT use a dictionary, internet, or other external help.\n\n"
    "Criterion: Words that begin with the letter F\n\n"
    "A human participant completed this task on Amazon Mechanical Turk.\n\n"
    "Here is their demographic and performance information:\n"
    f"Age: {age}\n"
    f"Highest degree: {education}\n"
    f"Number of correct responses: {num_correct}\n\n"
    "Now, please imagine that you are this participant. Your task is to generate as many words as possible that begin with the letter F, as if you were speaking aloud, and you have exactly one minute to do so. Respond in a way that reflects how this human participant might perform under this timed condition. "
    "Output only the words, one per line. Do not include any introductions, explanations, or extra text. If you add anything other than the words, you will be disqualified from the task."
)
    \end{lstlisting}
    \caption{Full prompt used to simulate the \textit{F} fluency task, including participant age, education, and number of correct responses, with instructions identical to those given to human participants}
    \label{fig:prompt-code}
\end{figure*}

\section{Model Selection Criteria}\label{app:model_selection}

Tables~\ref{tab:included} and~\ref{tab:excluded} summarize the performance of all 45 configurations in terms of adherence to the instructed number of responses.
Mean Absolute Error (MAE) was used to evaluate each configuration across all $106$ simulations, capturing the average magnitude of deviation between LLM-generated and human response counts regardless of direction:
\begin{equation} \text{MAE} = \frac{1}{N} \sum_{i=1}^{N} |y_i^{\text{LLM}} - y_i^{\text{Human}}| \end{equation}
\noindent where $N = 106$ is the number of participants and $y_i$ is the number of correct responses for participant $i$. Configurations with $\text{MAE} \leq 1.69$ (i.e., 10\% of the human mean of $16.89$) were considered to have successfully adhered to the performance constraints.

We additionally report Mean Bias Error (MBE), which captures the direction of deviation:
\begin{equation} \text{MBE} = \frac{1}{N} \sum_{i=1}^{N} (y_i^{\text{LLM}} - y_i^{\text{Human}}) \end{equation}
\noindent Positive MBE values indicate systematic overproduction relative to human participants, while negative values indicate underproduction. Because LLM-generated responses were subjected to the same error-checking criteria applied to human data---with repetitions being the most common error type when errors occurred, though overall error rates were very low---many models that closely followed the instructed number of responses tended to show slightly negative MBE values after error removal. 

Table~\ref{tab:included} also reports the number of outlier simulations per configuration, defined as cases where $|y_i^{\text{LLM}} - y_i^{\text{Human}}| > 5$.
Of the $33$ configurations meeting the MAE criterion, $21$ produced no outliers and were retained for participant-level and item-level analyses. The remaining $12$ configurations with outliers were excluded from these analyses due to unreliable individual simulations. However, for the LLM ensemble analysis, these $12$ configurations were included after removing their outlier simulations, in order to maximize the diversity of LLMs contributing to the ensemble.

\begin{table*}[htbp]
\centering
\caption{Model configurations with $\text{MAE} \leq 1.69$. Configurations with zero outliers were retained for participant-level and item-level analyses; those with outliers were included in ensemble analyses only (after removing outlier simulations).}
\label{tab:included}
\renewcommand{\arraystretch}{1.3}
\begin{tabular}{llrrr}
\hline
\textbf{Provider} & \textbf{Model} & \textbf{MAE} & \textbf{MBE} & \textbf{Outliers} \\
\hline
OpenAI
& GPT-5               & 0.00 &  0.00 & 0 \\
& GPT-5 mini                    & 0.00 &  0.00 & 0 \\
& GPT-5.2 (High Reasoning)    & 0.00 &  0.00 & 0 \\
& GPT-5.2 (Medium Reasoning)  & 0.00 &  0.00 & 0 \\
& GPT-5.2 (Low Reasoning)     & 0.00 &  0.00 & 0 \\
& o3-mini                   & 0.00 &  0.00 & 0 \\
& GPT-4.1                    & 0.06 & $-$0.06 & 0 \\
& GPT-3.5 Turbo              & 0.07 & $-$0.07 & 0 \\
& GPT-4o          & 0.16 & $-$0.16 & 0 \\
& GPT-5.2        & 1.50 &  1.46 & 2 \\
\hdashline
Anthropic
& Claude 4.5 Sonnet (Thinking) & 0.00 &  0.00 & 0 \\
& Claude 4 Sonnet (Thinking)   & 0.01 & $-$0.01 & 0 \\
& Claude 3.7 Sonnet (Thinking) & 0.02 & $-$0.02 & 0 \\
& Claude 4.5 Sonnet            & 0.06 & $-$0.06 & 0 \\
& Claude 3.7 Sonnet            & 0.12 & $-$0.08 & 0 \\
& Claude 4 Sonnet              & 0.14 &  0.05 & 0 \\
& Claude 4.6 Sonnet            & 0.75 &  0.71 & 3 \\
& Claude 4 Opus                & 0.84 & $-$0.07 & 4 \\
& Claude 4.1 Opus              & 1.06 &  0.62 & 6 \\
& Claude 4.5 Opus              & 1.43 &  1.30 & 10 \\
& Claude 4.6 Opus (Thinking)   & 1.48 &  1.46 & 9 \\
\hdashline
Google
& Gemini 2.5 Pro                       & 0.00 &  0.00 & 0 \\
& Gemini 3 Pro (High Thinking) & 0.00 &  0.00 & 0 \\
& Gemini 2.5 Flash                     & 0.04 &  0.04 & 0 \\
& Gemini 3 Pro (Low Thinking)  & 0.85 &  0.72 & 2 \\
& Gemini 3 Flash               & 0.86 & $-$0.86 & 8 \\
\hdashline
xAI
& Grok-4.1-fast (Reasoning) & 0.05 & $-$0.03 & 0 \\
& Grok-4-fast (Reasoning)   & 0.06 &  0.06 & 1 \\
\hdashline
Open-Source
& Kimi K2  & 0.03 & $-$0.03 & 0 \\
& Qwen3-2507          & 0.08 & $-$0.08 & 0 \\
& GLM-4.6             & 0.49 & $-$0.47 & 3 \\
& DeepSeek-V3.1       & 0.97 &  0.93 & 2 \\
& Llama 4 Maverick    & 1.29 &  1.27 & 9 \\
\hline
\end{tabular}
\vspace{0.5em}

\footnotesize \textbf{MAE} = Mean Absolute Error; \textbf{MBE} = Mean Bias Error (positive = overproduction, negative = underproduction); \textbf{Outliers} = number of participants with $|$LLM$-$Human$|$ $>$ 5.
\end{table*}

\begin{table}[htbp]
\centering
\caption{Model configurations with $\text{MAE} > 1.69$, excluded from all analyses.}
\label{tab:excluded}
\renewcommand{\arraystretch}{1.3}
\resizebox{\columnwidth}{!}{%
\begin{tabular}{lrr}
\hline
\textbf{Model} & \textbf{MAE} & \textbf{MBE} \\
\hline
\multicolumn{3}{l}{\textit{OpenAI}} \\
o4-mini          &  3.16 &  3.16 \\
GPT-5.1  &  3.91 &  3.85 \\
o3               & 35.39 & 34.67 \\
GPT-4 Turbo      & 57.17 & 57.17 \\
\hdashline
\multicolumn{3}{l}{\textit{Anthropic}} \\
Claude 4.5 Haiku             & 3.12 & 2.92 \\
Claude 4.6 Sonnet (Thinking) & 3.19 & 3.19 \\
Claude 4.6 Opus              & 3.87 & 3.81 \\
\hdashline
\multicolumn{3}{l}{\textit{xAI}} \\
Grok-4-fast   & 2.64 & 2.43 \\
Grok-4.1-fast  & 3.67 & 3.31 \\
\hdashline
\multicolumn{3}{l}{\textit{Open-Source Models}} \\
Cogito v2.1  &  3.20 &  2.16 \\
Mistral-Small-24B &  4.05 &  1.48 \\
Ministral-3-14B   & 18.62 & 18.62 \\
\hline
\end{tabular}}
\vspace{0.5em}

\footnotesize \textbf{MAE} = Mean Absolute Error; \textbf{MBE} = Mean Bias Error.
\end{table}

\section{Stepwise Regression Analyses of Linguistic Predictors of Production Frequency}\label{app:regression}

For human data, the stepwise regression yielded a significant model [$F(3,390)=86.85, p<.001$] that accounted for $40.0\%$ of the variance (adjusted $R^2=.396$). Word frequency emerged as the strongest predictor ($\beta =.34, t=6.04, p<.001$), followed by orthographic neighborhood size ($\beta=.254,t=5.91, p<.001$), and age of acquisition ($\beta=-.18,t=-3.17,p = .002$). This indicates that more frequent, orthographically dense, and earlier-acquired words were produced more often in the task.

For Claude 3.7 Sonnet, stepwise regression likewise produced a significant model [$F(3, 215) = 41.13, p<.001$] explaining $36.5\%$ of the variance in production frequency (adjusted $R^2=.356$). As with humans, word frequency was the strongest predictor ($\beta=.37,t = 5.16,p<.001$). However, unlike humans, orthographic neighborhood size was not retained in the model; instead, word length emerged as a significant predictor ($\beta=-.20,t=-3.56,p<.001$). Age of acquisition remained a significant predictor ($\beta=-.19, t=-2.59, p = .010$), consistent with the human pattern. o3-mini's regression model explained substantially less variance, only 18.8\% (adjusted $R^2=.179$) [$F(2,181)=20.94,p<.001$], retaining just two predictors: word frequency ($\beta=.33,t=4.59, p<.001$) and word length ($\beta=-.19, t=-2.60, p=.010$).

\section{Vocabulary Overlap Matrix}\label{app:overlap}
\begin{figure*}[h!] \centering \includegraphics[width=\textwidth]{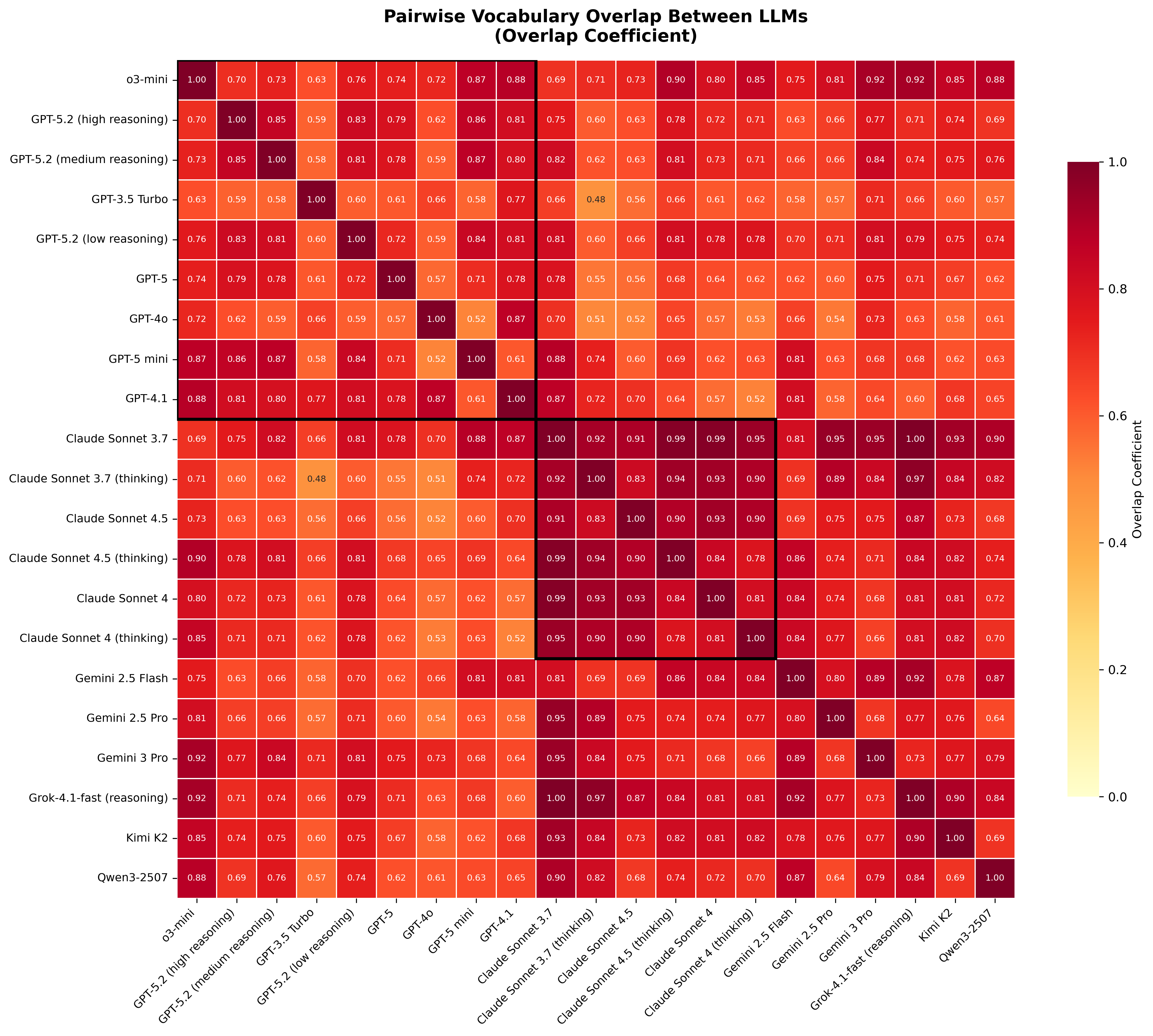} \caption{Pairwise vocabulary overlap coefficients between the $21$ LLM configurations. Higher values (darker) indicate greater shared vocabulary.} \label{fig:overlap} \end{figure*}

Figure~\ref{fig:overlap} presents the pairwise vocabulary overlap coefficients between all $21$ LLM configurations retained for participant-level analyses. The overlap coefficient between two vocabularies $A$ and $B$ is defined as:
\begin{equation} \text{Overlap}(A, B) = \frac{|A \cap B|}{\min(|A|, |B|)} \end{equation}
\noindent A value of $1.0$ indicates that the smaller vocabulary is entirely contained within the larger one, while lower values indicate greater lexical divergence.

\end{document}